\documentclass{article}
\usepackage[T1]{fontenc}
\usepackage{iclr2027_conference,times}

\usepackage{amsmath,amsfonts,bm}

\def\eqref#1{equation~\ref{#1}}

\def\1{\bm{1}}

\DeclareMathAlphabet{\mathsfit}{\encodingdefault}{\sfdefault}{m}{sl}
\SetMathAlphabet{\mathsfit}{bold}{\encodingdefault}{\sfdefault}{bx}{n}

\usepackage{amsmath,amssymb,algorithm,algorithmic,booktabs,multirow,graphicx}
\usepackage{microtype,xurl,hyperref,placeins}
\title{HarnessWAM: Bridging Prediction\\and Deliberation in World Action Models}
\author{
\textbf{Zhaopeng Gu}$^{1,2}$\,%
\textbf{Bingke Zhu}$^{1,2}$\thanks{Corresponding authors.}\,%
~\textbf{Tianxi Lin}$^{3,4}$\,%
\textbf{Guibo Zhu}$^{1,2}$\,%
\textbf{Yingying Chen}$^{1,2}$\,%
\textbf{Kai Wang}$^{1,3}$\\
\textbf{Tingyu Yuan}$^{1,2}$\,%
\textbf{Fan Yang}$^{1}$\,%
\textbf{Chaoyang Zhao}$^{1}$\,%
\textbf{Zhaowen Li}$^{3*}$\,%
\textbf{Peng Su}$^{3}$\,%
\textbf{Jinqiao Wang}$^{1,2,5}$\\
\normalfont $^{1}$ Institute of Automation, Chinese Academy of Sciences, Beijing, China\\
\normalfont $^{2}$ School of Artificial Intelligence, University of Chinese Academy of Sciences, Beijing, China\\
\normalfont $^{3}$ Yinwang Intelligent Technology Co., Ltd., Shenzhen, China\\
\normalfont $^{4}$ Beijing Institute of Technology, Beijing, China\\
\normalfont $^{5}$ Wuhan AI Research, Wuhan, China
}
\iclrfinalcopy
\hypersetup{pdftitle={HarnessWAM: Bridging Prediction and Deliberation in World Action Models},
  pdfauthor={Zhaopeng Gu, Bingke Zhu, Tianxi Lin, Guibo Zhu, Yingying Chen, Kai Wang, Tingyu Yuan, Fan Yang, Chaoyang Zhao, Zhaowen Li, Peng Su, Jinqiao Wang}}
\begin{document}
\maketitle
\lhead{HarnessWAM: Bridging Prediction and Deliberation in World Action Models}
\begin{abstract}
World Action Models~(WAMs) jointly learn environmental dynamics and robot actions, introducing priors over physical evolution into embodied control. However, finite-horizon prediction and action generation are insufficient for complex embodied tasks that require global planning, cross-stage state maintenance, execution verification, and failure recovery. We refer to this mismatch as the \emph{prediction-deliberation gap}. To address this gap, we propose \textbf{HarnessWAM}, an agentic framework for WAMs. HarnessWAM employs a vision-language-model-based Task Manager to maintain an evidence-grounded scene belief and organize the global instruction into a structured task graph. Capability-conditioned executable-space projection then compiles the graph into WAM-supported skill sequences that satisfy task dependencies and embodiment-state constraints. During execution, HarnessWAM couples a lightweight progress estimator with event-triggered Task Manager deliberation. The estimator supplies frequent execution cues, while the Task Manager uses observations, task state, and interaction history to verify outcomes and decide whether to advance, acquire evidence, revise the plan, or recover. Local recovery preserves acquired scene knowledge so that execution can resume after failure. HarnessWAM achieves state-of-the-art average performance on RoboMemArena, with full-task and subtask success rates of 64.4\% and 76.8\%, respectively, and reaches 23.7\% success on RoboCerebra Ideal. On a physical UR3 arm, full-task success reaches 95\% for shape-conditioned tabletop tidying and 90\% for memory-dependent box exploration. These results demonstrate the effectiveness of model-external agentic orchestration in bridging the prediction-deliberation gap for complex embodied tasks.
\end{abstract}
\section{Introduction}
\label{sec:introduction}
\begin{figure}[t]
\centering
\includegraphics[width=\linewidth]{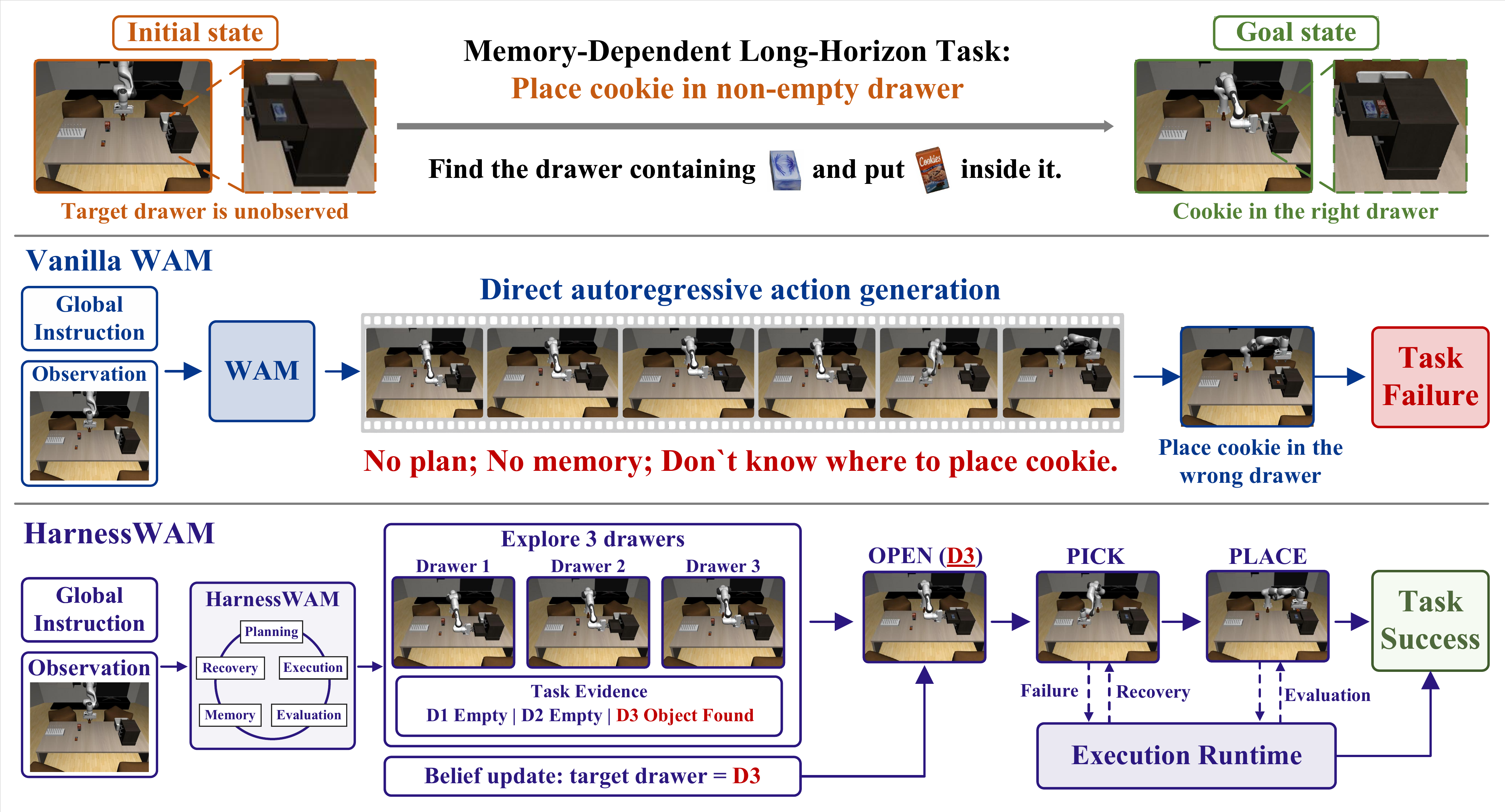}
\caption{The prediction-deliberation gap. Vanilla WAM denotes direct execution of the global instruction without external task state. HarnessWAM retains drawer observations, binds the target after occlusion, and coordinates skill execution, verification, and recovery.}
\label{fig:motivation}
\end{figure}

World Action Models (WAMs) jointly model future observations and robot actions, connecting predictions of physical dynamics to embodied control~\citep{li2026causal,ye2026world}. These learned dynamics support manipulation and generalization to unfamiliar tasks and environments~\citep{ye2026world}. However, complex tasks require decisions that extend beyond the model's prediction horizon. Placing a cookie in a non-empty drawer requires inspecting drawers, remembering their contents after closure, selecting the target, and verifying placement. A failed grasp further requires recovering the robot and resuming execution while retaining the evidence identifying the drawer. These requirements connect local action generation to persistent state, planning, verification, and recovery. We refer to this mismatch between predictive control and task-level deliberation as the \emph{prediction-deliberation gap}, illustrated in Figure~\ref{fig:motivation}.

Recent efforts address aspects of this gap through persistent visual memory~\citep{yang2026memorywam,wang2026dim} and hierarchical skill prediction~\citep{sun2026himem}. These mechanisms provide policies with historical context and temporal structure for skill execution. However, access to history and skill predictions does not ensure that target selection is grounded in observations, that operations satisfy their prerequisites, or that attempted actions achieve their intended effects. Language-model agents suggest a harness that maintains state and incorporates feedback across model calls~\citep{yao2022react,shinn2023reflexion,yang2024swe}. However, extending these harnesses to robots requires handling incomplete observations, physical constraints, and uncertain action outcomes. An embodied harness must therefore preserve task knowledge, ground semantic plans in supported skills, and verify physical effects before advancing execution.

To address this gap, we introduce \textbf{HarnessWAM}, an agentic framework for WAMs. HarnessWAM employs a vision-language-model (VLM) \emph{Task Manager} to maintain evidence-grounded scene belief and construct a task graph containing physical and cognitive operations. The graph includes evidence acquisition before unresolved targets are bound, allowing later skills to use targets identified through exploration. \emph{Capability-conditioned executable-space projection} then compiles the graph into validated WAM skills, checking dependencies against represented scene and robot constraints. Since admissible plans can still fail during execution, HarnessWAM couples a lightweight progress estimator with event-triggered semantic verification. Verification updates the task state and determines whether to advance, observe, revise, or recover. Revised suffixes undergo projection while executed history remains fixed. Recovery returns the robot toward its saved configuration while retaining evidence for subsequent decisions. The same task state therefore connects initial planning to verification, revision, and recovery throughout execution.

HarnessWAM achieves state-of-the-art performance on RoboMemArena, with full-task and subtask success rates of 64.4\% and 76.8\%, and reaches 23.7\% SR on RoboCerebra Ideal. Gains with FastWAM further demonstrate applicability across WAM backbones. On a UR3 arm, full-task success reaches 95\% for tabletop tidying and 90\% for box exploration.

Our main contributions are:
\begin{itemize}
    \item We identify the prediction-deliberation gap and introduce a WAM harness that connects local predictive control to persistent task-level deliberation.
    \item Evidence-grounded task graphs link target binding to information acquisition; capability-conditioned projection checks initial and revised plans against execution constraints.
    \item An event-driven runtime uses semantic verification to govern graph transitions, preserving executed history and acquired evidence across plan revision and recovery.
    \item State-of-the-art results across RoboMemArena, RoboCerebra Ideal, and UR3 experiments demonstrate the effectiveness of HarnessWAM.
\end{itemize}

\section{Related Work}
\label{sec:related_work}
\paragraph{World Models and World Action Models.}
World models learn environmental dynamics to predict how candidate actions change future observations. Visual foresight uses image predictions for model predictive control~\citep{ebert2017self,ebert2018visual}, while Dreamer learns latent dynamics and trains policies on imagined trajectories~\citep{hafner2025mastering}. Video-based policies transfer learned visual dynamics to action generation~\citep{du2023learning,wen2024vidman,liao2026genie}, while WAMs jointly learn these dynamics and robot actions~\citep{li2026causal,ye2026world,yuan2026fast}.

Recent WAMs extend this predictive foundation to longer interactions. MemoryWAM combines recent observations, anchor frames, and compressed history~\citep{yang2026memorywam}, while DiM-WAM uses event memory with progress supervision~\citep{wang2026dim}. HiMem-WAM couples hierarchical skill prediction with boundary-gated memory~\citep{sun2026himem}. AHA-WAM separates slower context updates from fast action generation~\citep{cai2026aha}. Together, these mechanisms extend the temporal context available for action generation and skill selection. However, this context alone does not ensure that an operation is executable or that its prerequisites have been achieved. HarnessWAM addresses this gap through evidence-grounded task graphs, capability-conditioned projection, and outcome verification that governs subsequent execution.

\paragraph{Harnesses for Embodied Agents.}
Robot agents ground language plans through skill affordances~\citep{ahn2022can} and geometric representations~\citep{rana2023sayplan,huang2023voxposer}, with task-and-motion planning formalizing physical feasibility~\citep{garrett2021integrated}. During execution, observed outcomes guide replanning~\citep{huang2022inner}. In concurrent work, Harness VLA combines such feedback with task-specific traces and global rules to coordinate frozen VLA skills and analytic primitives~\citep{zhang2026harness}. Complementary work addresses stage-aware progress estimation for policy training~\citep{chen2026sarm} and runtime failure detection~\citep{agia2024unpacking}. Recovery mechanisms then support continued execution after failure through local correction and additional attempts~\citep{ebert2018robustness,zhao2026flare}.

However, combining these mechanisms for WAM execution leaves a consistency problem when new evidence or failure changes the plan. Revised operations must respect acquired knowledge, skill preconditions, and completed execution history. HarnessWAM makes these dependencies explicit in an evidence-grounded task graph. Evidence acquisition resolves symbolic targets, capability-conditioned projection checks motor nodes against the validated WAM interface, and outcome verification governs dependent execution. When observations require replanning, suffix revision preserves executed node definitions and rechecks future operations. This coupling maintains consistency between task knowledge and execution constraints as the plan evolves.

\section{Method}
\label{sec:method}
\subsection{Problem Formulation}
\label{sec:problem_formulation}
\begin{figure}[t]
\centering
\includegraphics[width=\linewidth]{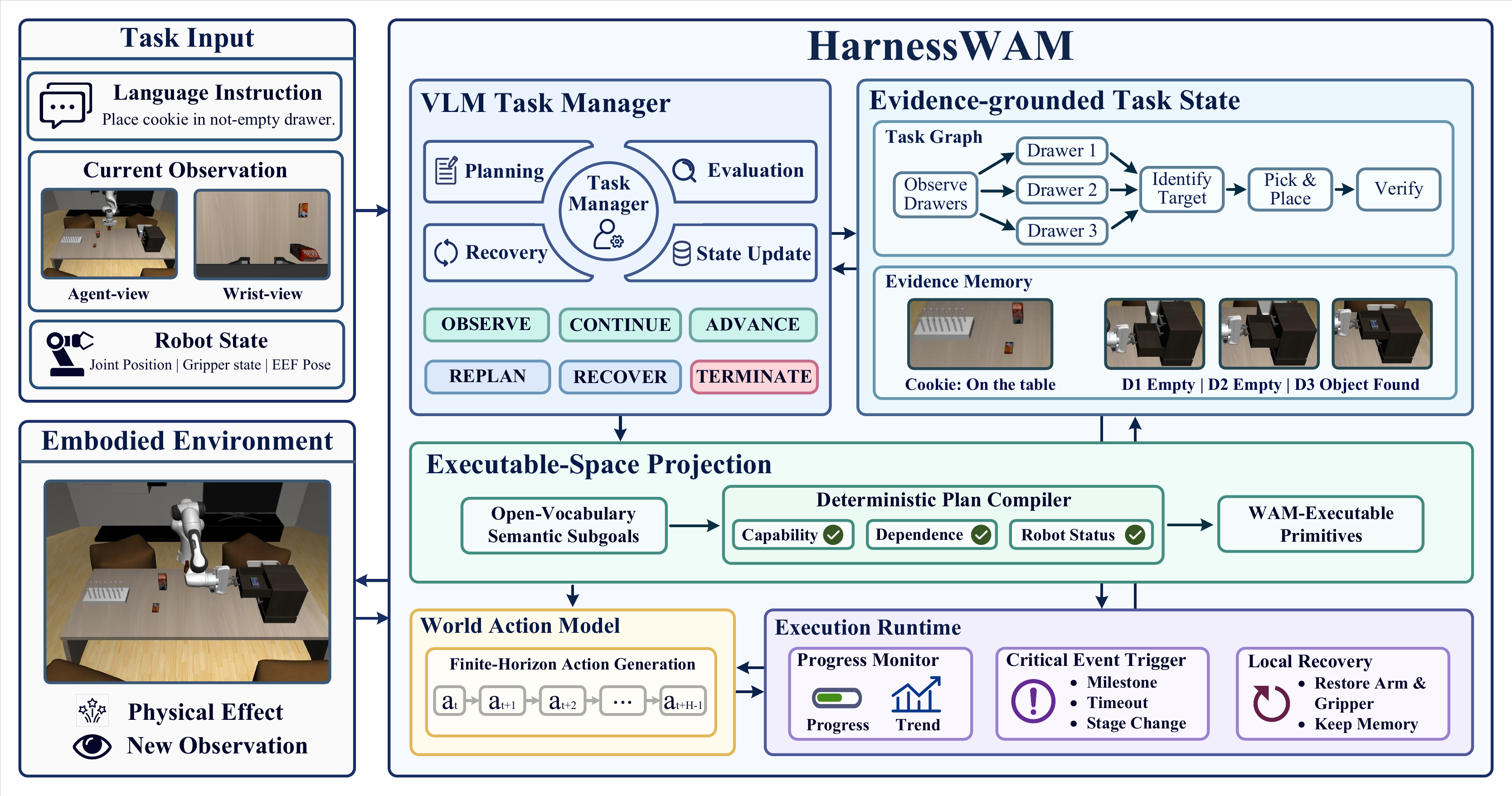}
\caption{HarnessWAM. The Task Manager maintains evidence-grounded state and a task graph. Capability projection compiles semantic goals into supported skills. Progress events connect the fast WAM loop to verification, future-plan revision, and local recovery.}
\label{fig:method_overview}
\end{figure}

Given a global instruction $x$, the robot receives dual-view RGB observations $o_t$ and proprioception $q_t$ at environment step $t$. A WAM $W_\theta$ produces an $H$-step action chunk $A_t=(a_t,\ldots,a_{t+H-1})$ conditioned on its available observation history and a local instruction $g_k$. The task-event index $k$ advances when execution calls for a task-level decision. To connect successive local skills, HarnessWAM maintains a runtime state $z_k=(B_k,G_k,M_k,r_k)$ and updates it at event time $\tau_k$:
\begin{equation}
 A_t=W_\theta(o_{\leq t},q_t,g_k),\qquad
 (z_{k+1},g_{k+1})=\mathcal H(x,z_k,o_{\tau_k},e_k),
 \label{eq:harness_transition}
\end{equation}
where $B_k$, $G_k$, and $M_k$ denote scene belief, task graph, and memory. The execution state $r_k$ tracks the active node, its attempt, and the remaining budget; $e_k$ is the triggering event. The harness selects a local instruction and updates task state from its outcome. Between events, the WAM generates action chunks for that instruction, as shown in Figure~\ref{fig:method_overview}.

\subsection{Evidence-Grounded Task State}
\label{sec:evidence_state}
Task-level decisions require both scene knowledge and the history of attempted operations. Scene belief $B_k$ records entities, attributes, and relations with confidence, supporting RGB evidence, and an epistemic status of observed, inferred, or unknown. Established facts, such as the contents of a closed drawer, remain available after occlusion. Task memory $M_k=(M_k^{\mathrm{task}},M_k^{\mathrm{evidence}})$ stores completed and failed nodes, retry counts, bindings, and revision history alongside salient RGB evidence. The Task Manager updates these records at event times, making earlier observations and outcomes available for subsequent decisions.

Using this state, the Task Manager encodes the instruction in a graph $G_k=(V_k,E_k,\mathcal X_k,\beta_k)$ of operations, dependencies, unresolved variables, and a partial binding. Each node is represented as
\begin{equation}
 v_i=(\mathrm{op}_i,\mathrm{arg}_i,\mathrm{pre}_i,\mathrm{eff}_i,\mathrm{term}_i,\mathrm{rec}_i),
 \label{eq:task_node}
\end{equation}
where the entries specify the operation, typed arguments, preconditions, expected effects, termination conditions, and recovery strategy, respectively. Motor nodes invoke physical skills; cognitive nodes acquire evidence, verify state, bind variables, or update memory. An unresolved target remains symbolic until evidence supports its assignment in $\beta_k$. Dependencies encode both physical and informational prerequisites, linking the order of motor operations to the evidence required for target selection and subsequent manipulation.

For example, a container-exploration task initially specifies a target through its unobserved contents. Inspection nodes make those contents observable, and a binding operation assigns the target variable from the accumulated evidence. Later motor nodes refer to that variable even after the containers close. Recording epistemic status distinguishes missing evidence from an observed negative result. Motor nodes await resolved arguments, completed dependencies, and satisfied preconditions, allowing evidence acquisition to precede target-dependent actions.

\subsection{Capability-Conditioned Executable-Space Projection}
\label{sec:executable_projection}
A task graph specifies semantic operations, but these operations may exceed the WAM's capabilities or violate physical prerequisites. We therefore restrict execution to a set $\mathcal P_W$ of validated WAM primitives, each specifying typed parameters, preconditions, effects, termination criteria, and recovery rules. Let $\mathcal L(\mathcal P_W)$ contain graphs whose motor nodes use these primitives, with cognitive nodes handled by the Task Manager. Let $\mathcal F(z_k)$ contain graphs consistent with scene state, bindings, dependencies, and robot configuration. For the proposed graph $G_k^{\mathrm{vlm}}$, we obtain
\begin{equation}
 G_k^{\mathrm{exec}}=
 \Pi_{\mathcal L(\mathcal P_W)\cap\mathcal F(z_k)}(G_k^{\mathrm{vlm}}),
 \label{eq:projection}
\end{equation}
where $\Pi$ denotes deterministic constrained compilation. The compiler checks argument types, dependencies, preconditions and effects, single-arm holding state, and graph acyclicity. It normalizes operator names and safe entity aliases into the WAM instruction vocabulary and expands supported compositions. For example, \texttt{PLACE(object, target)} requires the robot to have acquired the same object. If a graph is infeasible, the compiler reports the violated constraints to the Task Manager for replanning. Projection thus constrains both instruction wording and physical consistency; the WAM generates continuous trajectories for the compiled skills.

New evidence can resolve variables or invalidate operations. Suffix revision modifies the unexecuted portion of the graph through node insertion, deletion, replacement, reordering, or dependency updates, subject to the constraint
\begin{equation}
 G_{k+1}[V_k^{\mathrm{executed}}]=G_k[V_k^{\mathrm{executed}}],
 \label{eq:history_invariance}
\end{equation}
where $G[V]$ denotes the operation definitions and dependencies of nodes in $V$, and $V_k^{\mathrm{executed}}$ contains executed nodes. The recorded prefix preserves attempted operations and their dependencies, while updated belief and runtime status describe the current scene and execution outcome. If a placement attempt drops an object, the revised suffix must include reacquisition to satisfy the holding-state precondition of the next placement attempt. Every revised suffix undergoes projection before execution. Appendix~\ref{sec:revision_example} shows suffix revision after a placement failure.

The primitive vocabulary expands as additional WAM skills are validated; Appendix~\ref{sec:implementation_details} describes its ontology and scope. Projection establishes consistency under the represented constraints. However, satisfying these constraints does not ensure physical success, so execution requires outcome verification before subsequent nodes can proceed.

\subsection{Progress-Conditioned Event Control}
\label{sec:event_control}
To verify outcomes while maintaining continuous control, HarnessWAM uses a fast loop for local execution and progress estimation, with slower VLM deliberation triggered by execution events. In the fast loop, a prompt-conditioned estimator predicts continuous progress $p_t$, completion likelihood $c_t$, and a distribution $\pi_t^{\mathrm{bin}}$ over discrete progress levels:
\begin{equation}
 (p_t,c_t,\pi_t^{\mathrm{bin}})=F_\phi(o_{t-L+1:t},g_k),
 \label{eq:progress_estimator}
\end{equation}
where $L=5$ is the dual-view RGB window length. A frozen SigLIP2 encoder~\citep{tschannen2025siglip} and a four-layer causal Transformer provide spatial and temporal features. The estimator receives neither robot state nor action commands; instruction conditioning relates observed visual changes to the currently active skill. Training combines progress regression, progress-bin classification, temporal ranking, endpoint anchoring, completion prediction, and local monotonicity. Completion labels derive from temporal position in demonstrations, so the estimator supplies cues for verification without certifying physical success.

Progress milestones and skill-budget exhaustion trigger the slow Task Manager. Using current RGB observations, progress history, and runtime state, it verifies the active operation's outcome and updates scene belief and variable bindings. It classifies the outcome as successful, failed, or uncertain and selects whether to continue, advance, observe, replan, recover, or terminate execution. Advancement requires semantic outcome verification and the node-specific gripper condition; successors become eligible once all prerequisites are satisfied.

The Task Manager jointly returns an outcome, decision, state updates, and revision indicator:
\begin{equation}
 (\nu_k,d_k,\Delta B_k,\Delta\beta_k,\rho_k)
 =\mathcal T_{\mathrm{VLM}}(x,o_{\tau_k},B_k,G_k,M_k,r_k,\hat{s}^{\mathrm{prog}}_k,e_k),
 \label{eq:task_manager_main}
\end{equation}
where $\nu_k$ is the outcome, $d_k$ the decision, $\Delta B_k$ and $\Delta\beta_k$ the belief and binding updates, $\rho_k$ the suffix-revision indicator, and $\hat{s}^{\mathrm{prog}}_k$ the recent progress and completion trends.

An apparently complete motion can leave a required object relation unsatisfied. The Task Manager therefore checks the active node's expected effects against observations and retained evidence. Ambiguous imagery can trigger further observation before advancement or retry. These decisions update the belief, graph, and attempt history used for subsequent planning, connecting outcome verification to skill selection.

\subsection{Embodiment-State Recovery and Task Termination}
\label{sec:recovery}
When outcome verification indicates failure or uncertainty, the Task Manager uses visual evidence, progress, attempt history, and remaining budget to select a continuation. It can extend a slowly progressing attempt, obtain observations, revise an invalid strategy, or initiate recovery. To support recovery, the runtime records the arm and gripper configuration $(q_k^0,u_k^0)$ at the start of each motor node. Recovery applies multi-step joint control toward this configuration while retaining scene belief and interaction memory, then clears local WAM state. The Task Manager can revise the goal or future plan, which undergoes projection before execution resumes.

Objects may move during an unsuccessful attempt, so recovery does not restore an earlier world state. Subsequent decisions use updated observations together with retained evidence and the recorded outcomes of previous attempts.

The runtime declares completion when every required node is complete and final visual verification confirms the global goal. Execution terminates unsuccessfully when no feasible continuation remains or the task-level execution and recovery budget is exhausted. These limits bound physical interaction and recovery attempts. Appendix~\ref{sec:implementation_details} gives implementation details and pseudocode.

\section{Experiments}
\label{sec:experiments}
\subsection{Experimental Setup}
\label{sec:experimental_setup}
\paragraph{Tasks and protocol.}
RoboMemArena contains 26 tasks in transfer, occlusion, counting, and sequential execution, with 4, 11, 7, and 4 tasks, respectively; 68.9\% of subtasks depend on history~\citep{lei2026robomemarena}. Trajectories average 1,076 environment steps. Each WAM's configurations use 20 rollouts per task with matched initial states, random seeds, observations, and task-level execution budgets. RoboCerebra Ideal evaluates compositional manipulation in static, fully observable scenes~\citep{han2026robocerebra}, following its official protocol with 10 rollouts per task.

\begin{figure}[t]
    \centering
    \includegraphics[width=\linewidth]{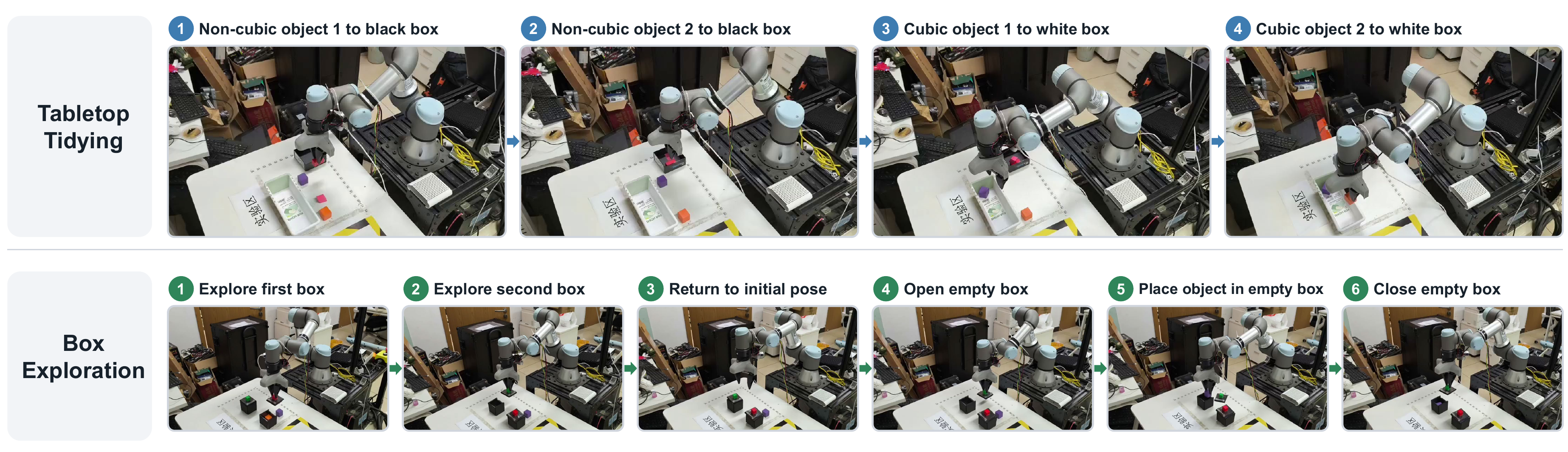}
    \caption{UR3 tasks. Tabletop Tidying sorts objects by shape. Box Exploration identifies an empty box through sequential inspection, then retains its identity for subsequent manipulation.}
    \label{fig:real_world_tasks}
\end{figure}

We evaluate two UR3 tasks, as shown in Figure~\ref{fig:real_world_tasks}. Tabletop Tidying assigns two cubic objects to a white box and two non-cubic objects to a black box, with four subtasks per trial. Box Exploration comprises six subtasks that inspect two boxes and place an object in the empty one, requiring retained evidence after the boxes close. We collect 100 demonstrations per task. All methods use these task-specific data with their respective checkpoints and are evaluated over 20 trials per task.

\paragraph{Metrics.}
For $N$ episodes, let $K_i$ count required subtasks in episode $i$ and $s_{ik}\in\{0,1\}$ indicate whether subtask $k$ satisfies its completion predicate. Full-task success and mean subtask completion are defined as
\begin{equation}
 \mathrm{SR}_{\mathrm{task}}=\frac1N\sum_{i=1}^N\prod_{k=1}^{K_i}s_{ik},\qquad
 \mathrm{SR}_{\mathrm{sub}}=\frac1N\sum_{i=1}^N\frac1{K_i}\sum_{k=1}^{K_i}s_{ik}.
 \label{eq:task_success_metric}
\end{equation}
Evaluation subtasks follow benchmark predicates independently of graph nodes. RoboMemArena metrics average rollouts within each task, then weight tasks equally. RoboCerebra SR measures the mean completion rate of key object-state transitions.

\paragraph{Models and comparisons.}
We fine-tune LingBot-VA~\citep{li2026causal} for each benchmark and additionally evaluate FastWAM~\citep{yuan2026fast} on RoboMemArena. For each WAM, whole-task and subtask checkpoints use identical training data and optimization settings. Subtask-based variants share the subtask checkpoint. LingBot-VA receives $256\times256$ agent- and wrist-view RGB and robot state. The Task Manager uses Qwen3-VL-32B-Instruct~\citep{bai2025qwen3} without task-specific fine-tuning or privileged visual inputs. Table~\ref{tab:robomemarena_main} uses the results for $\pi_{0.5}$~\citep{intelligence2025pi_}, HiF-VLA~\citep{lin2026hif}, MemoryVLA~\citep{shi2026memoryvla}, MemER~\citep{sridhar2026scaling}, and PrediMem from RoboMemArena~\citep{lei2026robomemarena}, and includes the concurrent Harness VLA framework~\citep{zhang2026harness}. RoboCerebra comparisons include $\pi_{0.5}$, OpenVLA~\citep{kim2024openvla}, GPT-4o Planner + OpenVLA, and HPE~\citep{han2026robocerebra}. UR3 comparisons use $\pi_{0.5}$, FastWAM, and LingBot-VA. Appendix~\ref{sec:additional_protocol} details these evaluation protocols.

\subsection{Main Results}
\label{sec:main_results}

\begin{table}[t]
    \caption{RoboMemArena success rates (\%). Standalone WAMs use whole-task checkpoints; HarnessWAM uses subtask checkpoints. Superscripts give absolute SR gains over the paired WAM, in the same units. Bold and underline denote the best and second-best results.}
    \label{tab:robomemarena_main}

    \begin{center}
    \setlength{\tabcolsep}{2.5pt}

    \begin{tabular*}{\linewidth}{@{\extracolsep{\fill}}lccccccccll@{}}
        \toprule
        \multicolumn{1}{c}{\multirow{2}{*}{\textbf{Method}}}
        & \multicolumn{2}{c}{\textbf{Transfer}}
        & \multicolumn{2}{c}{\textbf{Occlusion}}
        & \multicolumn{2}{c}{\textbf{Counting}}
        & \multicolumn{2}{c}{\textbf{Sequence}}
        & \multicolumn{2}{c}{\textbf{Average}} \\
        \cmidrule(lr){2-3}\cmidrule(lr){4-5}\cmidrule(lr){6-7}\cmidrule(lr){8-9}\cmidrule(lr){10-11}
        & $\mathrm{SR}_{\mathrm{task}}$ & $\mathrm{SR}_{\mathrm{sub}}$
        & $\mathrm{SR}_{\mathrm{task}}$ & $\mathrm{SR}_{\mathrm{sub}}$
        & $\mathrm{SR}_{\mathrm{task}}$ & $\mathrm{SR}_{\mathrm{sub}}$
        & $\mathrm{SR}_{\mathrm{task}}$ & $\mathrm{SR}_{\mathrm{sub}}$
        & $\mathrm{SR}_{\mathrm{task}}$ & $\mathrm{SR}_{\mathrm{sub}}$ \\
        \midrule
        $\pi_{0.5}$ & 20.0 & 42.8 & 12.7 & 17.2 & 14.3 & 50.9 & 60.0 & 71.6 & 21.5 & 38.7 \\
        HiF-VLA & 17.5 & 38.9 & 12.7 & 27.1 & 8.6 & 45.9 & 42.5 & 70.2 & 16.9 & 39.8 \\
        MemoryVLA & 15.0 & 37.2 & 7.3 & 13.1 & 14.3 & 55.1 & 37.5 & 65.2 & 15.0 & 35.3 \\
        MemER & 20.0 & 36.1 & 16.4 & 33.2 & 27.1 & 65.1 & 65.0 & 79.1 & 27.3 & 49.1 \\
        PrediMem & \underline{22.5} & \underline{45.2} & 27.3 & 38.4 & 45.7 & 69.3 & 72.5 & \underline{89.5} & 38.5 & 55.2 \\
        Harness VLA & \textbf{50.0} & \textbf{58.3} & 0.0 & 0.0 & 12.1 & 50.0 & \textbf{82.5} & 84.5 & 23.6 & 35.4 \\
        \midrule
        FastWAM & 1.3 & 5.4 & 12.3 & 24.7 & 11.4 & 28.3 & 27.5 & 55.8 & 12.7 & 27.5 \\
        + HarnessWAM & 8.8 & 21.3 & 36.4 & 53.7 & \underline{55.7} & \underline{73.0} & 45.0 & 61.7 & 38.7\textsuperscript{+26.0} & 55.1\textsuperscript{+27.6} \\

        \midrule
        LingBot-VA & 11.3 & 12.5 & \underline{63.2} & \underline{66.6} & 37.1 & 68.2 & 68.8 & 74.8 & \underline{49.0} & \underline{60.0} \\
        + HarnessWAM & \underline{22.5} & 26.5 & \textbf{76.8} & \textbf{87.7} & \textbf{59.3} & \textbf{80.6} & \underline{81.3} & \textbf{90.6} & \textbf{64.4}\textsuperscript{+15.4} & \textbf{76.8}\textsuperscript{+16.8} \\

        \bottomrule
    \end{tabular*}
    \end{center}
\end{table}

\paragraph{Simulation.}
On RoboMemArena, HarnessWAM achieves state-of-the-art performance using LingBot-VA, as shown in Table~\ref{tab:robomemarena_main}. The full-task and subtask SRs reach 64.4\% and 76.8\%, respectively, compared with 38.5\% and 55.2\% for PrediMem.

With LingBot-VA, HarnessWAM yields absolute gains of 15.4\% in full-task SR and 16.8\% in subtask SR over whole-task execution. The gains reach 26.0\% and 27.6\% with FastWAM. Combining subtask-trained control with HarnessWAM improves completion in every task family for both backbones.

On RoboCerebra Ideal, HarnessWAM achieves 23.70\% SR, compared with 21.92\% for GPT-4o Planner + OpenVLA and 21.10\% for HPE, as shown in Table~\ref{tab:robocerebra_main}.

\begin{table}[htbp]
    \caption{Success rate on RoboCerebra Ideal (\%).}
    \label{tab:robocerebra_main}
\begin{center}

    \setlength{\tabcolsep}{6pt}
    \begin{tabular}{lc}
        \toprule
        Method & Ideal SR $\uparrow$ \\
        \midrule
        $\pi_{0.5}$                  & 1.88 \\
        OpenVLA (fine-tuned)         & 7.84 \\
        GPT-4o Planner + OpenVLA     & \underline{21.92} \\
        HPE Framework                & 21.10 \\
        \midrule
        \textbf{HarnessWAM}          & \textbf{23.70} \\
        \bottomrule
    \end{tabular}
\end{center}
\end{table}

\paragraph{Physical manipulation.}
In Table~\ref{tab:real_world_results}, HarnessWAM reaches 95.0\% full-task SR and 96.3\% subtask SR for tidying, compared with 30.0\% and 55.0\% for LingBot-VA. For exploration, the corresponding rates are 90.0\% and 95.8\%, compared with 5.0\% and 25.8\%. Neither $\pi_{0.5}$ nor FastWAM completes an exploration trial under the shared training data. HarnessWAM thus supports repeated rule-based manipulation and subsequent actions conditioned on earlier observations.

\begin{table}[htbp]
    \caption{Real-world success rates (\%) with successful/total counts in parentheses. Each method is evaluated over 20 trials per task. Bold and underline denote the best and second-best results.}
    \label{tab:real_world_results}
\begin{center}

    \setlength{\tabcolsep}{3pt}
    \begin{tabular*}{\linewidth}{@{\extracolsep{\fill}}lcccc@{}}
        \toprule
        \multirow{2}{*}{Method}
        & \multicolumn{2}{c}{Tabletop Tidying}
        & \multicolumn{2}{c}{Box Exploration} \\
        \cmidrule(lr){2-3}\cmidrule(lr){4-5}
        & $\mathrm{SR}_{\mathrm{task}}$ & $\mathrm{SR}_{\mathrm{sub}}$
        & $\mathrm{SR}_{\mathrm{task}}$ & $\mathrm{SR}_{\mathrm{sub}}$ \\
        \midrule
        $\pi_{0.5}$ & 5.0 (1/20) & 31.3 (25/80) & 0.0 (0/20) & 0.0 (0/120) \\
        FastWAM & 25.0 (5/20) & 40.0 (32/80) & 0.0 (0/20) & 2.5 (3/120) \\
        LingBot-VA & \underline{30.0 (6/20)} & \underline{55.0 (44/80)} & \underline{5.0 (1/20)} & \underline{25.8 (31/120)} \\
        \textbf{HarnessWAM} & \textbf{95.0 (19/20)} & \textbf{96.3 (77/80)} & \textbf{90.0 (18/20)} & \textbf{95.8 (115/120)} \\
        \bottomrule
    \end{tabular*}
\end{center}
\end{table}

\subsection{Ablation Study}
\label{sec:ablations}
\paragraph{Architectural controls.}
Table~\ref{tab:ablations} compares orchestration strategies sharing a subtask-trained LingBot-VA checkpoint. WAM + Static Plan fixes the initial skill sequence without memory updates, reaching 52.5\% full-task SR. This sequence cannot respond to new evidence. Free-form orchestration permits online adaptation through prompts generated every 50 steps from current RGB and recent decisions. However, these updates lack structured task state and executable constraints, yielding 11.9\% full-task SR. HarnessWAM couples state updates with constrained plan revision to incorporate new evidence into supported skill sequences, reaching 64.4\% full-task SR.

\begin{table}[htbp]
    \caption{Architectural controls and ablations on RoboMemArena (\%). All settings share the same subtask-trained LingBot-VA checkpoint. Bold and underline mark the best and second-best results.}
    \label{tab:ablations}
\begin{center}

    \setlength{\tabcolsep}{1.3pt}

    \begin{tabular*}{\linewidth}{@{\extracolsep{\fill}}lcccccccccc@{}}
        \toprule
        \multicolumn{1}{c}{\multirow{2}{*}{\textbf{Method}}}
        & \multicolumn{2}{c}{\textbf{Transfer}}
        & \multicolumn{2}{c}{\textbf{Occlusion}}
        & \multicolumn{2}{c}{\textbf{Counting}}
        & \multicolumn{2}{c}{\textbf{Sequence}}
        & \multicolumn{2}{c}{\textbf{Average}} \\
        \cmidrule(lr){2-3}\cmidrule(lr){4-5}\cmidrule(lr){6-7}\cmidrule(lr){8-9}\cmidrule(lr){10-11}
        & $\mathrm{SR}_{\mathrm{task}}$ & $\mathrm{SR}_{\mathrm{sub}}$
        & $\mathrm{SR}_{\mathrm{task}}$ & $\mathrm{SR}_{\mathrm{sub}}$
        & $\mathrm{SR}_{\mathrm{task}}$ & $\mathrm{SR}_{\mathrm{sub}}$
        & $\mathrm{SR}_{\mathrm{task}}$ & $\mathrm{SR}_{\mathrm{sub}}$
        & $\mathrm{SR}_{\mathrm{task}}$ & $\mathrm{SR}_{\mathrm{sub}}$ \\

        \midrule
        \multicolumn{11}{l}{\textit{(a) Architectural controls}} \\
        VLM + WAM (free-form)        & 0.0 & 3.9 & 23.2 & 44.3 & 7.9 & 28.8 & 0.0 & 16.9 & 11.9 & 29.7 \\
        WAM + Static Plan           & 15.0 & 16.7 & 60.0 & 73.6 & \underline{52.9} & \textbf{82.5} & 68.8 & 89.6 & 52.5 & 69.7 \\
        \midrule
        \multicolumn{11}{l}{\textit{(b) Planning and task state}} \\
        w/o task state              & 15.0 & 17.2 & 60.0 & 72.9 & 52.1 & 79.5 & 68.8 & 89.6 & 52.3 & 68.7 \\
        w/o executable projection   & 5.0 & 10.6 & 30.9 & 53.1 & 9.3 & 47.2 & 43.8 & 59.4 & 23.1 & 45.9 \\
        w/o suffix revision         & 20.0 & 23.5 & 72.7 & \textbf{88.4} & 50.7 & 77.9 & 62.5 & 76.0 & 57.1 & 73.7 \\
        \midrule
        \multicolumn{11}{l}{\textit{(c) Execution control}} \\
        w/o progress events       & 13.8 & 17.5 & 59.1 & 66.9 & 27.1 & 63.3 & 55.0 & 67.3 & 42.9 & 58.4 \\
        Progress-only switching   & \textbf{23.8} & \underline{27.5} & \underline{75.0} & 85.5 & 51.4 & 77.4 & 67.5 & \textbf{91.0} & \underline{59.6} & \underline{75.2} \\
        w/o recovery              & \underline{22.5} & \textbf{28.5} & 73.2 & 85.0 & 51.4 & 79.3 & \underline{70.0} & 83.5 & 59.0 & 74.5 \\
        \midrule
        HarnessWAM                & \underline{22.5} & 26.5 & \textbf{76.8} & \underline{87.7} & \textbf{59.3} & \underline{80.6} & \textbf{81.3} & \underline{90.6} & \textbf{64.4} & \textbf{76.8} \\
        \bottomrule
    \end{tabular*}
\end{center}
\end{table}

\paragraph{State and projection.}
Removing structured scene facts, history, and bindings reduces average full-task SR to 52.3\%, with the largest decline on occlusion. Under occlusion, successful execution depends on evidence retained from earlier observations. Separately, disabling projection reduces full-task SR to 23.1\% despite retaining task state and feedback. Access to task knowledge is therefore insufficient when proposed operations lack capability, argument, and dependency checks. Persistent evidence and executable constraints serve complementary roles in connecting task reasoning to WAM control under changing observations and physical conditions.

\paragraph{Online graph revision.}
We freeze future nodes and dependencies while reusing the full system's initial plans and retaining memory, binding, verification, and recovery. On occlusion tasks, subtask SR remains close to HarnessWAM at 88.4\% versus 87.7\%, while full-task SR falls from 76.8\% to 72.7\%. Memory and binding updates retain newly acquired evidence, while the fixed suffix leaves subsequent operations unchanged. Online revision adapts these operations and their dependencies to the updated task state, allowing new observations to guide the remaining execution.

\paragraph{Execution control.}
Progress events align Task Manager calls with skill execution; replacing them with periodic calls reduces full-task SR from 64.4\% to 42.9\%. At each intervention, the Task Manager checks the active skill's effects before authorizing advancement. Without semantic verification and gripper-state gating, progress-only switching reaches 75.2\% subtask SR and 59.6\% full-task SR, as estimated completion may precede satisfaction of the next skill's prerequisites. When verification detects failure, recovery enables another attempt. Without recovery, failed or exhausted attempts terminate execution, lowering full-task SR to 59.0\%.

\paragraph{Plan quality.}
Table~\ref{tab:projection_analysis} compares raw and projected VLM plans. Projection raises executability from 13.8\% to 72.9\%, dependency satisfaction from 58.1\% to 92.9\%, and binding accuracy from 21.3\% to 88.3\%. Syntactically valid plans can still select incorrect targets or violate skill prerequisites. Projection jointly checks bindings, dependencies, and physical constraints before WAM execution.

\begin{table}[htbp]
    \caption{Plan quality before and after executable-space projection (\%).}
    \label{tab:projection_analysis}
\begin{center}

    \setlength{\tabcolsep}{4pt}
    \begin{tabular}{lcccc}
        \toprule
        Plan representation & Syntax $\uparrow$ & Dependencies $\uparrow$ & Binding $\uparrow$ & Executability $\uparrow$ \\
        \midrule
        Raw VLM plan                    & \underline{60.8} & \underline{58.1} & \underline{21.3} & \underline{13.8} \\
        Projected plan                  & \textbf{95.2} & \textbf{92.9} & \textbf{88.3} & \textbf{72.9} \\
        \bottomrule
    \end{tabular}
\end{center}
\end{table}

\subsection{Qualitative Results}
\label{sec:qualitative_analysis}
\begin{figure}[htbp]
    \centering
    \includegraphics[width=\linewidth]{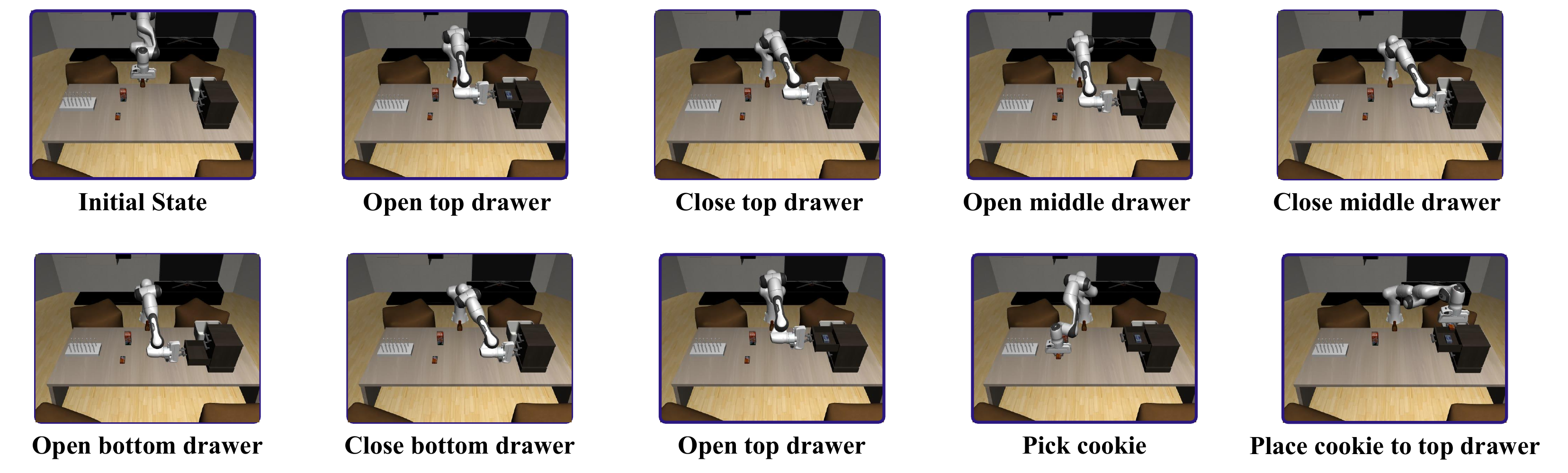}
    \caption{A RoboMemArena Task~4 rollout. Sequential drawer inspection supplies evidence for delayed target binding and subsequent object placement after the target contents become occluded.}
    \label{fig:task4_rollout}
\end{figure}

Figure~\ref{fig:task4_rollout} shows a RoboMemArena Task~4 rollout. HarnessWAM opens and closes three drawers, retaining evidence about their contents. Once the drawers close, their contents are occluded, and the current observation alone no longer identifies the required destination. The Task Manager uses the retained evidence to bind the target variable to the non-empty top drawer and specify subsequent actions. This binding carries the inspection result into the placement instruction, allowing the WAM to execute a local skill with a resolved target. The robot then reopens this drawer and places the object inside. This trajectory instantiates the evidence acquisition and target binding illustrated in Figure~\ref{fig:motivation}, linking retained observations to subsequent manipulation.

\FloatBarrier

\section{Conclusion}
\label{sec:conclusion}
HarnessWAM connects finite-horizon WAM control to persistent task-level deliberation. Evidence-grounded task graphs and capability-conditioned projection organize local skills, while event-triggered verification determines how observed outcomes change subsequent execution. Plan revision and recovery preserve the evidence and history needed to continue after failure. Results on RoboMemArena, RoboCerebra Ideal, and two UR3 tasks support this framework across history-dependent and compositional manipulation. Ablations identify contributions from planning, task state, verification, and recovery, while plan diagnostics show how constraint checks improve represented consistency. Performance remains limited by the WAM's skill repertoire and the accuracy of VLM state estimation. These limitations motivate broader skill coverage and calibrated uncertainty estimates to guide task-level deliberation under partial observability and uncertain action outcomes.

\label{sec:main_end}
\bibliography{references}
\bibliographystyle{iclr2027_conference}
\clearpage
\appendix
\section{Representation and Implementation Details}
\label{sec:implementation_details}
\subsection{Scene Facts, Memory, and Graph Nodes}
Given a global instruction, the Task Manager organizes execution into motor and cognitive nodes. A motor node instantiates a supported primitive with task-specific arguments, such as an object to acquire or a container to open. Cognitive nodes obtain and process the evidence needed to identify these arguments and assess the effects of motor operations. Both types of node use a persistent scene belief to retain information across observations.

The scene belief records each fact as
\begin{equation}
    f=(s,p,o,v,\eta,c,\mathcal E),
    \label{eq:scene_fact}
\end{equation}
where $s$, $p$, $o$, and $v$ denote the subject, predicate, object, and value. The remaining fields record how the fact is known: $\eta$ takes the value \texttt{observed}, \texttt{inferred}, or \texttt{unknown}, $c$ records confidence, and $\mathcal E$ references supporting RGB evidence. This distinction separates an unobserved container from one observed to be empty. Once its contents have been observed, closing the container does not by itself invalidate the recorded fact.

HarnessWAM records the interactions associated with these facts in a two-part memory:
\begin{equation}
    M_k=(M_k^{\mathrm{task}},M_k^{\mathrm{evidence}}),
    \label{eq:memory}
\end{equation}
where $M_k^{\mathrm{task}}$ stores completed and failed nodes, retry counts, variable bindings, and plan revisions, while $M_k^{\mathrm{evidence}}$ stores visual evidence from salient task events. At an event time $\tau_k$, the Task Manager combines this history with the current observation to update the scene belief and memory:
\begin{equation}
    (B_{k+1},M_{k+1})
    =\mathcal U_{\mathrm{VLM}}
    (x,o_{\tau_k},B_k,M_k,e_k).
    \label{eq:belief_update}
\end{equation}
The updated state retains the facts and outcomes needed for subsequent decisions, together with references to their visual evidence. The Task Manager can therefore use information acquired earlier in the interaction even when the corresponding scene content is no longer visible.

The Task Manager organizes operations over this state in a directed task graph
\begin{equation}
    G_k=(V_k,E_k,\mathcal X_k,\beta_k),
    \label{eq:task_graph}
\end{equation}
where $V_k$ contains motor and cognitive nodes, $E_k$ records their dependencies, and $\mathcal X_k$ contains unresolved variables. Each node follows Equation~\ref{eq:task_node}, which specifies its operation, arguments, preconditions, effects, termination conditions, and recovery strategy. Dependencies require preceding actions or evidence-acquisition steps to be completed before a node becomes eligible for execution.

The partial binding $\beta_k:\mathcal X_k\rightharpoonup\mathcal O$ assigns resolved variables to entities in the scene-entity set $\mathcal O$. If the initial observation does not identify a required target, the graph retains a symbolic variable and schedules the observations needed to resolve it. Cognitive nodes then verify the observed state, update memory, and bind the variable when sufficient evidence becomes available. Subsequent motor nodes use the resulting binding, allowing target-dependent manipulation to follow exploration even after the target is occluded.

\subsection{Primitive Ontology}
The graph's motor nodes require an operation vocabulary with typed arguments and physical semantics. We organize this vocabulary into an extensible ontology:
\begin{equation}
    \mathcal P^\star
    =
    \mathcal P_{\mathrm{motion}}
    \cup\mathcal P_{\mathrm{grasp}}
    \cup\mathcal P_{\mathrm{contact}}
    \cup\mathcal P_{\mathrm{articulation}}
    \cup\mathcal P_{\mathrm{assembly}}
    \cup\mathcal P_{\mathrm{tool}},
    \label{eq:primitive_ontology}
\end{equation}
where the six sets correspond to free-space motion, object acquisition and release, contact-rich interaction, articulated-object manipulation, assembly, and tool use, respectively. Within each family, a primitive specifies the physical operation and accepts object or target arguments. The WAM realizes an instantiated operation by generating a trajectory from the current observation. The interface for each primitive is
\begin{equation}
    p=
    \bigl(
    \tau_p,\Theta_p,
    \operatorname{Pre}_p,
    \operatorname{Eff}_p,
    \operatorname{Term}_p,
    \operatorname{Rec}_p
    \bigr),
    \label{eq:primitive}
\end{equation}
where $\tau_p$ denotes the interaction type, $\Theta_p$ is a typed parameter space, and the remaining terms specify preconditions, expected effects, termination criteria, and recovery rules.

A compact operation vocabulary is consistent with the recurrent and low-dimensional structure observed in motor behavior. In a study of two housekeepers and two machinists, the ten most frequent grasp types account for 81\% of observed grasp duration and 72\% of grasp instances~\citep{bullock2013grasp}. Kinematic studies also identify low-dimensional structure within movement: two motion primitives explain approximately 95\% of cumulative joint-trajectory variance in human reaching~\citep{moro2012kinematic}. For hand postures formed as if grasping 57 familiar objects, the first two principal components explain more than 80\% of the variance across 15 measured joint angles~\citep{santello1998postural}. At the level of physical interaction, robotic manipulation can be organized into a finite set of relative-motion classes between rigid bodies~\citep{morrow1997manipulation}. These observations motivate parameterized operation families whose physical semantics remain explicit while trajectory generation is delegated to the WAM.

For a particular WAM, the executable primitive set is determined by the skills that the model has acquired and that have been empirically validated:
\begin{equation}
    \mathcal P_W
    =
    \left\{
    p\in\mathcal P^\star
    \mid
    p\ \text{has a validated realization under}\ W_\theta
    \right\}.
    \label{eq:wam_primitive_set}
\end{equation}
The implemented motor vocabulary comprises \texttt{PICK}, \texttt{PLACE}, \texttt{MOVE\_TO}, \texttt{POUR}, \texttt{OPEN}, and \texttt{CLOSE}. Each operation has a canonical WAM prompt grammar and typed object or target arguments. During compilation, the arguments and operation sequence determine the required holding-state transitions. Placement requires the matching object to be held, while opening or closing an articulated object requires an empty gripper. The compiler checks these requirements together with graph dependencies before admitting the plan for execution. Additional operation families enter the executable vocabulary as their WAM realizations are validated.

\subsection{Evidence-Driven Plan Revision}
\label{sec:revision_example}
\begin{figure}[t]
    \centering
    \includegraphics[width=\linewidth]{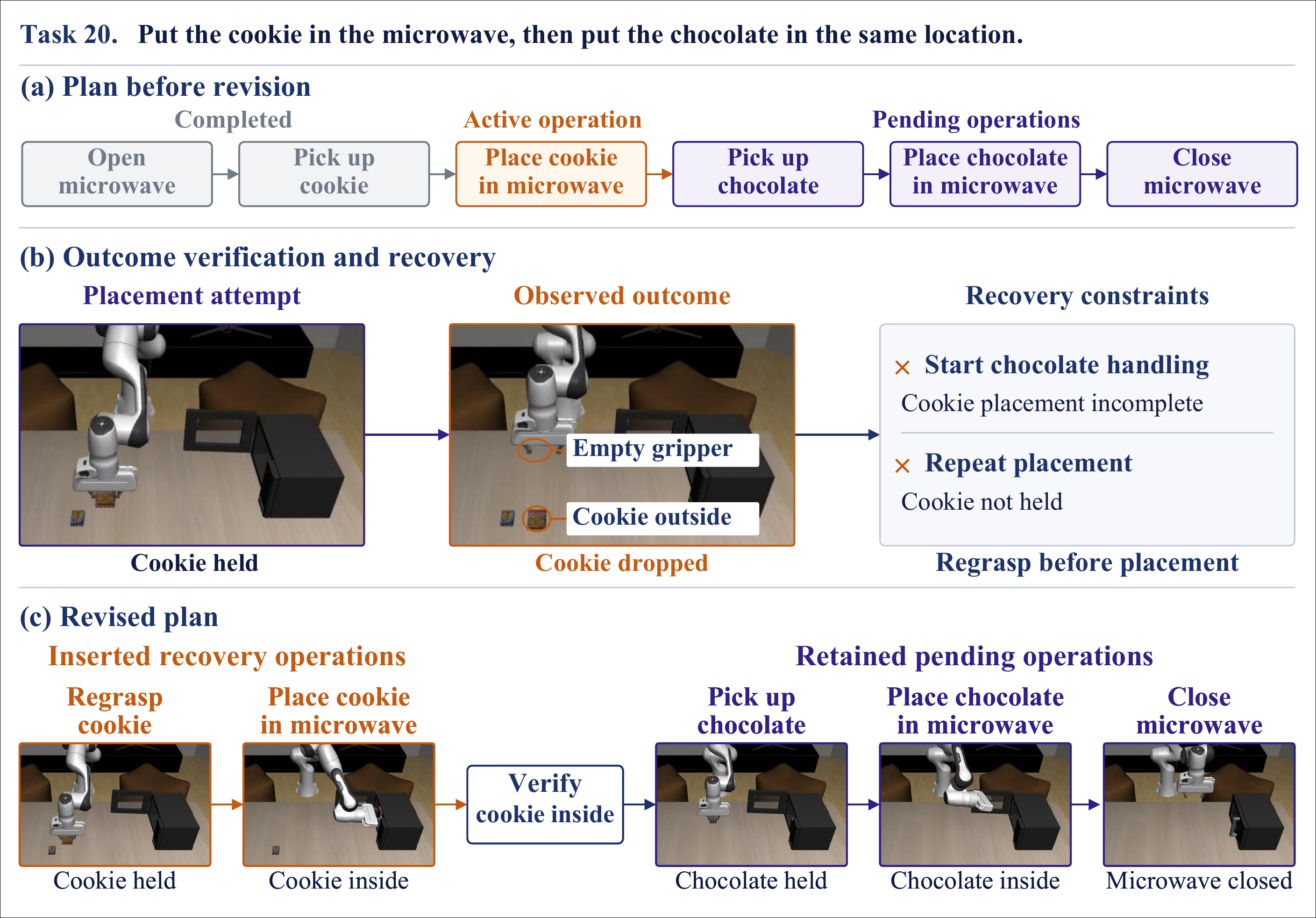}
    \caption{Evidence-driven plan revision in RoboMemArena Task 20. (a) The plan separates completed operations, the active placement attempt, and pending operations. (b) The dropped cookie and empty gripper leave the placement incomplete and prevent an immediate placement retry. (c) The revised suffix adds regrasping and placement before the retained chocolate and closing operations. Outcome verification gates progression to chocolate handling.}
    \label{fig:plan_revision}
\end{figure}

Figure~\ref{fig:plan_revision} presents a rollout from RoboMemArena Task 20, which requires placing a cookie in the microwave, followed by chocolate in the same container. The initial plan opens the microwave, schedules acquisition and placement for each object, and closes the door. After opening and cookie acquisition have completed, the first placement attempt drops the cookie onto the table. Outcome verification identifies the cookie outside the microwave and an empty gripper.

The incomplete placement leaves the dependency for chocolate handling unsatisfied. The empty gripper also violates the holding-state precondition for another placement attempt. The Task Manager therefore revises the unexecuted suffix by inserting a \texttt{PICK} operation to reacquire the cookie and a new \texttt{PLACE} attempt targeting the same microwave. The pending chocolate and closing operations remain after these recovery nodes.

Executable-space projection checks the revised sequence against the WAM's supported primitives, task dependencies, and holding-state requirements. Regrasping restores the condition required for placement. Once the Task Manager verifies that the cookie is inside the microwave, execution proceeds to acquire the chocolate, place it in the microwave, and close the door. Completed operations and the failed attempt remain in memory, while the executed graph prefix retains its operation definitions and dependencies.

\subsection{Progress Estimation}
\label{sec:progress_training}
The fast execution loop estimates visual progress to determine when the Task Manager should assess the active operation. The estimator takes a local window of dual-view RGB observations and the active subtask instruction, with no robot states or actions as inputs. A frozen \textit{SigLIP2-base-patch16-256} encoder extracts image patch features, global image features, and an instruction embedding. Prompt-conditioned attention uses this embedding to pool spatial evidence from each camera.

The pooled features are fused across views together with their changes relative to the first frame in the local window and the instruction embedding. A four-layer causal Transformer processes the resulting sequence, and separate output heads predict continuous progress, progress-bin probabilities, and completion likelihood. Training updates the pooling, fusion, temporal, and output layers while keeping the encoder frozen.

\paragraph{Supervision and sampling.}
Each training sample contains two local windows from the same annotated subtask segment and shares a single subtask instruction. The segment boundaries provide temporal supervision for both windows. For segment $i$ with inclusive start frame $s_i$ and exclusive end frame $e_i$, normalized progress is
\begin{equation}
 y_i(f)=\frac{f-s_i}{\max(1,e_i-s_i-1)},\qquad s_i\leq f<e_i,
 \label{eq:progress_target}
\end{equation}
where $f$ is the frame index. Each window contains $L=5$ consecutive frames ending at a sampled anchor, and its targets follow their normalized temporal positions within the demonstration. Indices before the segment start repeat the first frame so that windows retain a fixed length near the boundary.

Three sampling modes cover different temporal relationships. Random pairs use separated anchors subject to minimum frame and progress gaps. Endpoint pairs place anchors near the beginning and end of the segment. Local pairs restrict candidate anchors to nearby frames while maintaining a minimum progress gap. If repeated random sampling fails to find an admissible pair, the sampler uses the segment's first and last frames as anchors.

Let $q\in\{a,b\}$ index the earlier and later windows, respectively. At each valid frame $t$, the model predicts progress $p_{it}^q$, a bin distribution $\pi_{it}^q$, and completion probability $c_{it}^q$. We write the progress prediction and target at the final frame as $\hat p_i^q$ and $y_i^q$. These frame-level and window-level outputs are trained jointly with
\begin{equation}
    \mathcal L_{\mathrm{prog}}
    =
    \lambda_r\mathcal L_{\mathrm{reg}}
    +\lambda_b\mathcal L_{\mathrm{bin}}
    +\lambda_{\mathrm{rank}}\mathcal L_{\mathrm{rank}}
    +\lambda_e\mathcal L_{\mathrm{endpoint}}
    +\lambda_s\mathcal L_{\mathrm{success}}
    +\lambda_m\mathcal L_{\mathrm{mono}},
    \label{eq:progress_loss}
\end{equation}
where the six terms supervise continuous progress, discrete progress levels, temporal ordering, segment boundaries, completion cues, and local monotonicity, respectively. Their coefficients specify the loss weights listed in Table~\ref{tab:training_configuration}. A term evaluates to zero when its supervision set is empty.

\paragraph{Continuous progress regression.}
The regression term supervises the final frame of each window, matching the prediction used during online execution. It uses the smooth $L_1$ function $h(u)=u^2/2$ for $|u|<1$ and $h(u)=|u|-1/2$ otherwise:
\begin{equation}
 \mathcal L_{\mathrm{reg}}=
 \frac{\sum_i w_i\bigl[h(\hat p_i^a-y_i^a)+h(\hat p_i^b-y_i^b)\bigr]}
 {\max(1,\sum_i w_i)},
 \label{eq:progress_regression}
\end{equation}
where $w_i$ is the annotation confidence for the segment boundaries, clipped to $[0.1,1]$ and set to one when unspecified. Since the temporal targets depend on these boundaries, confidence weighting reduces the contribution of less certain annotations. The two window errors anchor progress predictions at both sampled positions within the segment.

\paragraph{Progress-bin classification.}
The auxiliary classification head assigns each valid frame to one of $K$ discretized progress levels. Its objective is
\begin{equation}
 \mathcal L_{\mathrm{bin}}=
 -\sum_{q\in\{a,b\}}\frac{1}{|\mathcal V_q|}
 \sum_{(i,t)\in\mathcal V_q}\log\pi_{it}^q\!\left(b_{it}^q\right),
 \qquad b_{it}^q=\operatorname{round}\!\left((K-1)y_{it}^q\right),
 \label{eq:progress_bins}
\end{equation}
where $\mathcal V_q$ contains the valid frames in windows of type $q$, and $y_{it}^q$ is the corresponding normalized temporal target. With $K=11$, the class centers cover zero through one at intervals of 0.1. Cross-entropy supplies coarse progress supervision throughout each window, complementing regression at its final frame.

\paragraph{Temporal ranking.}
The ranking loss enforces a margin between progress predictions at the final frames of the two windows:
\begin{equation}
 \mathcal L_{\mathrm{rank}}=
 \frac{1}{|\mathcal R|}\sum_{i\in\mathcal R}
 \left[m-\left(\hat p_i^b-\hat p_i^a\right)\right]_+,
 \label{eq:progress_ranking}
\end{equation}
where $[u]_+=\max(0,u)$, $m$ is the ranking margin, and $\mathcal R$ contains pairs whose later anchor has strictly larger temporal position and target progress. This term penalizes reversed or insufficiently separated predictions, with degenerate pairs excluded from supervision.

\paragraph{Endpoint anchoring.}
Endpoint-pair sampling introduces an additional regression objective near the segment boundaries:
\begin{equation}
 \mathcal L_{\mathrm{endpoint}}=
 \frac{1}{2|\mathcal E|}\sum_{i\in\mathcal E}
 \sum_{q\in\{a,b\}}h(\hat p_i^q-y_i^q),
 \label{eq:progress_endpoints}
\end{equation}
where $\mathcal E$ denotes the endpoint-pair subset of the batch. The earlier anchor is $\min(s_i+L-1,e_i-1)$, and the later anchor is $e_i-1$. Applying Equation~\ref{eq:progress_target} to these anchors gives the two targets, including a potentially nonzero target at the end of the initial window. The additional regression term increases the weight of observations near the segment boundaries.

\paragraph{Completion prediction.}
Completion supervision derives weak labels from normalized temporal position. Frames at or below $\tau_-=0.70$ are negative, and frames at or above $\tau_+=0.95$ are positive. The intermediate interval is excluded because its temporal position does not provide a definite completion label. The resulting masked binary cross-entropy is
\begin{equation}
 \mathcal L_{\mathrm{success}}=-\frac{1}{|\mathcal S|}
 \sum_{j\in\mathcal S}\left[\alpha z_j\log c_j+(1-z_j)\log(1-c_j)\right],
 \label{eq:progress_completion}
\end{equation}
where $\mathcal S$ contains labeled valid frames from both windows, $z_j=\mathbf 1[y_j\geq\tau_+]$, and $c_j$ is the predicted completion probability. To account for the relative frequencies of labeled frames, the positive-class weight is $\alpha=\operatorname{clip}(N_-/\max(1,N_+),1,\alpha_{\max})$, where $N_-$ and $N_+$ count negative and positive labels in the batch and $\alpha_{\max}=5$. Because supervision derives from demonstration timing, the completion head supplies a cue for deliberation. The Task Manager subsequently checks the operation's physical effects before authorizing advancement.

\paragraph{Local monotonicity.}
Within each window, a monotonicity penalty discourages decreases in progress between adjacent frames:
\begin{equation}
 \mathcal L_{\mathrm{mono}}=
 \frac{1}{2}\sum_{q\in\{a,b\}}\frac{1}{|\mathcal A_q|}
 \sum_{(i,t)\in\mathcal A_q}[p_{it}^q-p_{i,t+1}^q]_+,
 \label{eq:progress_monotonicity}
\end{equation}
where $\mathcal A_q$ contains adjacent valid-frame pairs. Penalizing decreases encourages consistency with the temporal order of the demonstration. This constraint acts within a window and complements the ranking loss between window endpoints. Both are soft penalties, so predicted progress can still change in response to visual evidence.

\paragraph{Optimization and validation.}
Episodes are assigned to disjoint training and validation sets before subtask sampling. On the training split, AdamW optimizes the trainable layers with gradient-norm clipping and a cosine learning-rate schedule following linear warmup. Validation windows use fixed anchors at normalized positions 0, 0.25, 0.5, 0.75, and 1, with consecutive anchors paired across the segment. Progress predictions are compared with the temporal targets over valid validation-window frames, and the checkpoint with the lowest mean absolute error is selected.
\subsection{Training Configuration}
\label{sec:training_configuration}
Table~\ref{tab:training_configuration} lists the settings for progress-estimator training and LingBot-VA fine-tuning. The estimator is trained for 100 epochs with eight window pairs per device in each minibatch. Each pair contains two windows from one subtask segment, as specified in Section~\ref{sec:progress_training}.

The whole-task and subtask checkpoints for each WAM use the same training data and optimization settings, with supervision aligned to complete tasks or individual subtasks.

LingBot-VA fine-tuning uses one training sample per device and accumulates gradients over ten minibatches before each optimizer update. The optimization settings are shared across RoboMemArena, RoboCerebra Ideal, and UR3 training. In each setting, the Task Manager retains its pretrained parameters throughout WAM fine-tuning and progress-estimator training.

FastWAM is trained on RoboMemArena for 50,000 steps with AdamW, a batch size of 64, and weight decay of $1\times10^{-2}$. Its learning rate decays from $1\times10^{-4}$ to $1\times10^{-6}$ following a cosine schedule.

\begin{table}[htbp]
\caption{Key training hyperparameters. Loss weights are listed in the order of Equation~\ref{eq:progress_loss}. Batch sizes are specified per device.}
\label{tab:training_configuration}
\begin{center}
\begin{tabular}{p{0.40\linewidth}p{0.52\linewidth}}
\toprule
Parameter & Setting \\
\midrule
\multicolumn{2}{l}{\textbf{(a) Progress estimator}}\\
Encoder & SigLIP2-base-patch16-256, frozen \\
Temporal window & 5 consecutive frames \\
Temporal Transformer & 4 layers, width 512, 8 attention heads \\
Training and validation split & 90\% and 10\% of episodes \\
Loss weights & $(1.0,0.5,0.5,0.2,0.25,0.05)$ \\
Ranking margin & 0.05 \\
Optimizer & AdamW \\
Peak learning rate & $1\times10^{-4}$ \\
Weight decay & $1\times10^{-2}$ \\
Batch size & 8 window pairs \\
Training duration & 100 epochs \\
Learning-rate schedule & Cosine decay with 5\% linear warmup \\
\midrule
\multicolumn{2}{l}{\textbf{(b) LingBot-VA fine-tuning}}\\
Optimizer & AdamW \\
Learning rate & $1\times10^{-5}$ \\
Weight decay & 0.1 \\
Batch size & 1 training sample \\
Gradient accumulation & 10 minibatches per optimizer update \\
Training budget & 50,000 optimizer updates \\
Learning-rate schedule & Constant after 10 warmup updates \\
Gradient-norm clipping & 2.0 \\
\bottomrule
\end{tabular}
\end{center}
\end{table}

\subsection{Task Manager Outputs}
When progress milestones or skill-budget exhaustion trigger deliberation, the Task Manager evaluates the active operation using current RGB, recent progress estimates, and persistent runtime state. It jointly returns an outcome, an execution decision, and the associated state changes:
\begin{equation}
\begin{aligned}
y_k&=(\nu_k,d_k,\Delta B_k,\Delta\beta_k,\rho_k)\\
&=\mathcal T_{\mathrm{VLM}}(x,o_{\tau_k},B_k,G_k,M_k,r_k,\hat{s}^{\mathrm{prog}}_k,e_k),
\end{aligned}
\label{eq:task_manager_decision}
\end{equation}
where $\hat{s}^{\mathrm{prog}}_k$ summarizes recent progress, completion likelihood, and temporal trends. The outcome $\nu_k$ classifies the active operation as \texttt{success}, \texttt{failure}, or \texttt{uncertain}. Using this assessment and the runtime state, the Task Manager selects $d_k$ from \texttt{continue}, \texttt{advance}, \texttt{observe}, \texttt{replan}, \texttt{recover}, and \texttt{terminate}.

The accompanying changes $\Delta B_k$ and $\Delta\beta_k$ update scene belief and variable bindings. If these updates require changes to future operations, $\rho_k$ signals suffix revision, after which the revised graph undergoes capability and feasibility checks. An \texttt{advance} decision marks the active node complete when semantic verification and its gripper condition are satisfied. A successor becomes eligible when its arguments are bound, its predecessors are complete, and its preconditions hold.

\subsection{Inference Procedure}
Algorithm~\ref{alg:harnesswam} begins by initializing task state and projecting the initial VLM plan. The scheduler selects a node whose prerequisites are satisfied. A cognitive node acquires or processes evidence and updates task state before selection resumes. A motor node records the arm and gripper configuration, then executes WAM action chunks under progress monitoring. Triggered deliberation determines whether to continue the operation, advance, gather evidence, revise the plan, recover, or terminate.

The runtime state $r$ tracks the active node, attempt status, and remaining budget. Motor execution, active observation, and recovery consume environment steps from the task budget $\Omega$. Execution succeeds when all required nodes are complete and final visual verification confirms the global goal. It fails if the budget is exhausted or no feasible continuation remains.
\begin{algorithm}[t]
\caption{Event-driven HarnessWAM inference}
\label{alg:harnesswam}
\begin{algorithmic}[1]
\REQUIRE Instruction $x$, initial RGB observation $o_0$, WAM $W_\theta$, task budget $\Omega$
\ENSURE Task outcome in $\{\texttt{success},\texttt{failure}\}$
\STATE $t\leftarrow0$; $(B,M,r)\leftarrow\textsc{Initialize}(x,o_0,\Omega)$
\STATE $G^{\mathrm{vlm}}\leftarrow\textsc{Plan}(x,B,M)$
\STATE $G\leftarrow\textsc{Project}(G^{\mathrm{vlm}},W_\theta,B)$
\WHILE{$\Omega>0$ and $G$ contains an unfinished required node}
    \STATE $v\leftarrow\textsc{SelectReadyNode}(G)$; fail if no feasible continuation remains
    \IF{$v$ is a cognitive node}
        \STATE Acquire evidence and update $B$, $M$, bindings, and the graph suffix
        \STATE Mark $v$ complete if its termination condition is satisfied
        \STATE Reproject any revised suffix; update execution state and budget
        \STATE \textbf{continue}
    \ENDIF
    \STATE $(q^0,u^0)\leftarrow$ current arm and gripper state; activate $v$
    \WHILE{$v$ is active and $\Omega>0$}
        \STATE $A_t\leftarrow W_\theta(o_{\leq t},q_t,\textsc{Prompt}(v))$
        \STATE Execute $A_t$ within budget; advance $t$ and refresh $o_t$, $q_t$, $r$, and $\Omega$
        \STATE $\hat{s}^{\mathrm{prog}}\leftarrow F_\phi(\text{recent RGB},\textsc{Prompt}(v))$
        \STATE $e\leftarrow\textsc{Event}(\hat{s}^{\mathrm{prog}},r,G)$
        \IF{$e$ triggers deliberation}
            \STATE $(\nu,d,\Delta B,\Delta\beta,\rho)\leftarrow
            \mathcal T_{\mathrm{VLM}}(x,o_t,B,G,M,r,\hat{s}^{\mathrm{prog}},e)$
            \STATE Update $B$, $M$, and bindings; if $\rho$, revise and reproject the suffix
            \IF{$d=\texttt{advance}$ and the node-specific gripper condition holds}
                \STATE Mark $v$ complete; \textbf{break}
            \ELSIF{$d=\texttt{observe}$}
                \STATE Acquire evidence, update task state and budget, and reproject any revision
            \ELSIF{$d=\texttt{replan}$}
                \STATE Revise and reproject the graph suffix; deactivate $v$
            \ELSIF{$d=\texttt{recover}$}
                \STATE Move toward $(q^0,u^0)$, update budget, and clear local WAM state
                \STATE Revise and project the recovery plan; deactivate $v$
            \ELSIF{$d=\texttt{terminate}$}
                \RETURN \texttt{failure}
            \ENDIF
        \ENDIF
    \ENDWHILE
\ENDWHILE
\IF{all required nodes are complete and $\textsc{VerifyGoal}(x,o_t,B)$}
    \RETURN \texttt{success}
\ELSE
    \RETURN \texttt{failure}
\ENDIF
\end{algorithmic}
\end{algorithm}
\section{Additional Evaluation Protocol Details}
\label{sec:additional_protocol}
\subsection{Shared Evaluation Protocol}
The simulation evaluations follow the settings in Section~\ref{sec:experimental_setup}. On RoboMemArena, the full system and the variants in Table~\ref{tab:ablations} cover all 26 tasks with 20 rollouts per task. These settings use the same seeds, and corresponding task and seed pairs have matched initial states. RoboCerebra Ideal uses the official evaluation settings with 10 rollouts per task.

Within each benchmark, standalone LingBot-VA uses its whole-task checkpoint, while HarnessWAM and the settings in Table~\ref{tab:ablations} share its subtask checkpoint. Both checkpoints use the same training data and optimization settings and receive $256\times256$ agent-view and wrist-view RGB together with robot state. The Task Manager uses Qwen3-VL-32B-Instruct without task-specific fine-tuning. It reasons from RGB observations without depth, segmentation labels, or privileged simulator state.

\subsection{FastWAM Configuration}
\label{sec:fastwam_configuration}
The two FastWAM settings in Table~\ref{tab:robomemarena_main} use checkpoints trained for their respective execution interfaces. Direct execution uses a FastWAM checkpoint trained on complete tasks and conditioned on the global instruction. HarnessWAM uses a separate checkpoint trained to execute individual subtasks, with local instructions supplied by the Task Manager. Both checkpoints use the same training data and FastWAM optimization settings in Appendix~\ref{sec:training_configuration}. Each setting evaluates all 26 tasks with 20 rollouts per task under matched initial states, random seeds, and execution budgets.

\subsection{Harness VLA Configuration}
\label{sec:harness_vla_configuration}
The Harness VLA baseline in Table~\ref{tab:robomemarena_main} retains the original framework~\citep{zhang2026harness}. An LLM agent coordinates a frozen $\pi_0$-family VLA with analytic motion and gripper primitives. The agent uses these primitives to prepare non-contact motions and invokes the VLA for contact-rich manipulation. Task-specific memory and global memory retain execution strategies across attempts. For each task, exploration operates on a single reference configuration with a budget of 20 complete-task episodes. Evaluation then comprises 20 rollouts per task using the accumulated memories.

\subsection{Architectural Controls and Ablations}
Table~\ref{tab:robomemarena_main} reports standalone WAMs trained for whole-task execution. The architectural controls and component ablations in Table~\ref{tab:ablations} share HarnessWAM's subtask-trained LingBot-VA checkpoint.

\paragraph{Standalone WAMs.}
The WAM uses its whole-task checkpoint and receives the complete task instruction throughout the rollout, with no subtask instructions during training or evaluation. It generates each action chunk under this fixed language condition. The controller uses neither an initial VLM-generated subtask sequence nor online VLM prompt selection.

\paragraph{WAM + Static Plan.}
The VLM analyzes the initial scene and generates a linear sequence of WAM prompts. This sequence remains fixed during execution. The controller executes the sequence without online plan revision, milestone memory updates, VLM failure diagnosis, or final VLM verification.

\paragraph{VLM + WAM (free-form).}
At initialization and every 50 environment steps, the VLM receives the global instruction, dual-view RGB, active WAM prompt, remaining execution budget, and a text transcript of up to 16 recent decisions. From these inputs, it chooses to continue, switch prompts, retry, finish, or fail. A prompt switch changes the WAM's local instruction. A retry clears the local execution segment and restarts the current prompt from the robot's current state, without restoring a saved robot configuration.

The controller maintains its interaction history through the decision transcript and generates unrestricted prompts. It has no structured task graph, evidence-grounded scene memory, explicit variable binding, prompt normalization, or executable projection. Its periodic VLM decisions also operate without a learned progress estimator or semantic node verification.

\paragraph{Without task state.}
The Task Manager receives the current imagery and global instruction without stored scene facts, interaction history, or variable bindings. Consequently, earlier visual evidence is unavailable through task state after the corresponding objects or relations become occluded.

\paragraph{Without executable projection.}
The VLM-generated plan bypasses projection and its checks on supported operations, typed arguments, preconditions, holding state, and dependencies. Execution retains task-state updates, progress events, VLM semantic verification, gripper-state gating, suffix revision, and recovery.

\paragraph{Without suffix revision.}
Each task and seed reuses the initial plan from the corresponding full HarnessWAM rollout. The controller keeps future node definitions and dependency edges fixed by disabling insertion, deletion, replacement, reordering, and dependency modification. Existing symbolic variables can still be bound to scene entities, and scene memory continues to incorporate new observations. These state updates leave the graph structure unchanged. Progress events, semantic verification, failure diagnosis, and embodiment-state recovery remain active during execution.

\paragraph{Without progress events.}
The learned progress estimator is disabled, and the Task Manager is invoked every 50 environment steps. At each scheduled call, it evaluates semantic completion from the current evidence. A successful assessment authorizes advancement only when the active node's gripper condition is also satisfied.

\paragraph{Progress-only switching.}
The controller advances between subtasks using the learned progress and completion predictions. VLM semantic outcome verification and gripper-state gating are disabled, so transitions proceed without checking the operation's observed effects.

\paragraph{Without recovery.}
The controller retains progress estimation and semantic outcome verification but sets the retry budget to zero. A detected failure or exhausted skill budget therefore terminates execution without restoring the saved robot configuration or initiating another attempt.

\subsection{Plan-Quality Evaluation}
Each VLM plan is generated once and saved for evaluation before and after executable-space projection. The compiler maps operation names and safe entity aliases to the canonical execution vocabulary and checks capability, dependency, and physical constraints. Table~\ref{tab:projection_analysis} reports syntax, dependencies, binding, and executability for the raw and projected plans.

\paragraph{Syntax.}
Syntax measures the percentage of nodes that satisfy the node schema and operator-specific prompt grammar. The schema requires unique node identifiers and list-valued dependencies. For motor nodes, the prompt must be accepted by the corresponding WAM operator; cognitive nodes have an empty motor prompt.

\paragraph{Dependencies.}
Dependency satisfaction measures whether the predicted graph preserves the required ordering of motor operations. A longest common subsequence aligns predicted operations with a reference operation sequence. For each adjacent reference pair, the aligned predecessor must be an ancestor of the aligned successor in the predicted graph. The score is the fraction of reference relations represented by these paths, with missing aligned operations counted as unsatisfied relations. Graphs with duplicate identifiers, missing dependency targets, or cycles receive a score of zero.

\paragraph{Binding.}
Binding accuracy uses the same operation alignment to compare predicted object and target arguments with their reference entities. It measures the fraction of matching reference argument slots, counting missing operations or arguments as incorrect. For projected plans, the comparison uses the canonical arguments returned by the compiler.

\paragraph{Executability.}
Executability requires a non-empty plan with valid syntax and graph structure, all required dependency relations, consistent holding-state transitions, and agreement with the reference motor-prompt sequence. A plan passes only when these conditions hold jointly; the executability score is the percentage of plans that pass.

Holding-state validation follows acquired and released objects through the operation sequence. Transport and placement must refer to the object currently held, and opening or closing an articulated object requires an empty gripper. Pouring additionally requires a preceding move to the matching target. Each operation is checked against the holding state left by preceding actions.

For plans with deferred decisions, unresolved variables and explicit observation or binding nodes can leave future motor operations unspecified. In this case, sequence matching accepts a prefix of the reference sequence. This prefix must satisfy all syntax, dependency, and holding-state checks.

\subsection{Physical Evaluation}
The physical evaluations use a UR3 arm for Tabletop Tidying and Box Exploration, with task sequences shown in Figure~\ref{fig:real_world_tasks}. For each task, 100 demonstration episodes provide the training data shared by the four methods in Table~\ref{tab:real_world_results}. Each method uses its respective model checkpoint and is evaluated in 20 trials per task.

\paragraph{Tabletop Tidying.}
The scene contains two cubic and two non-cubic objects, a white box, and a black box. The task assigns cubic objects to the white box and non-cubic objects to the black box. Its four subtasks transfer the two non-cubic objects first, followed by the two cubic objects, with one object transferred per subtask. Each transfer is complete when the designated object is placed in the box specified by its shape, and full-task success requires all four transfers to succeed.

\paragraph{Box Exploration.}
The scene contains two black boxes, one initially empty, and an object on the tabletop. The robot must inspect both boxes to identify the empty one before placing the object inside it. The first two subtasks inspect the boxes in sequence, with opening and closing each box constituting one subtask. The third subtask returns the arm to its initial pose. The remaining three subtasks reopen the previously identified empty box, place the tabletop object inside, and close it. Because both boxes are closed after inspection, selecting the correct box requires retaining the earlier observation. Full-task success requires all six subtasks to succeed.

\FloatBarrier

\end{document}